\documentclass[10pt,twocolumn,letterpaper]{article}

\usepackage{wacv}              

\usepackage{graphicx}
\usepackage{amsmath}
\usepackage{amssymb}
\usepackage{booktabs}
\usepackage[acronym]{glossaries}
\usepackage{subcaption}
\usepackage{placeins}

\usepackage[pagebackref,breaklinks,colorlinks]{hyperref}

\usepackage[capitalize]{cleveref}
\crefname{section}{Sec.}{Secs.}
\Crefname{section}{Section}{Sections}
\Crefname{table}{Table}{Tables}
\crefname{table}{Tab.}{Tabs.}

\def\wacvPaperID{2154} 
\def\confName{WACV}
\def\confYear{2025}

\newacronym{ipm}{IPM}{Inverse Perspective Mapping}
\newacronym{bev}{BEV}{Bird's-Eye-View}
\newacronym{svs}{SVS}{Surround-View System}
\newacronym{iri}{IRI}{International Roughness Index}
\newacronym{bfgs}{BFGS}{Broyden–Fletcher–Goldfarb–Shanno}
\newacronym{mde}{MDE}{Mean Distance Error}
\newacronym{adas}{ADAS}{Advanced Driver Assistance System}
\newacronym{slam}{SLAM}{Simultaneous Localization and Mapping}

\begin{document}

\title{Click-Calib: A Robust Extrinsic Calibration Method for Surround-View Systems}

\author{Lihao Wang
\thanks{Valeo, San Mateo, USA. lihao.wang@valeo.com}
}

\maketitle

\begin{abstract}
   \gls{svs} is an essential component in \gls{adas} and requires precise calibrations. However, conventional offline extrinsic calibration methods are cumbersome and time-consuming as they rely heavily on physical patterns. Additionally, these methods primarily focus on short-range areas surrounding the vehicle, resulting in lower calibration quality in more distant zones. To address these limitations, we propose Click-Calib, a pattern-free approach for offline \gls{svs} extrinsic calibration. Without requiring any special setup, the user only needs to click a few keypoints on the ground in natural scenes. Unlike other offline calibration approaches, Click-Calib optimizes camera poses over a wide range by minimizing reprojection distance errors of keypoints, thereby achieving accurate calibrations at both short and long distances. Furthermore, Click-Calib supports both single-frame and multiple-frame modes, with the latter offering even better results. Evaluations on our in-house dataset and the public WoodScape dataset demonstrate its superior accuracy and robustness compared to baseline methods. Code is available at \url{https://github.com/lwangvaleo/click_calib}.
\end{abstract}

\section{Introduction}
\label{sec:intro}

\begin{figure}[thpb]
  \centering
   \includegraphics[width=0.9\linewidth]{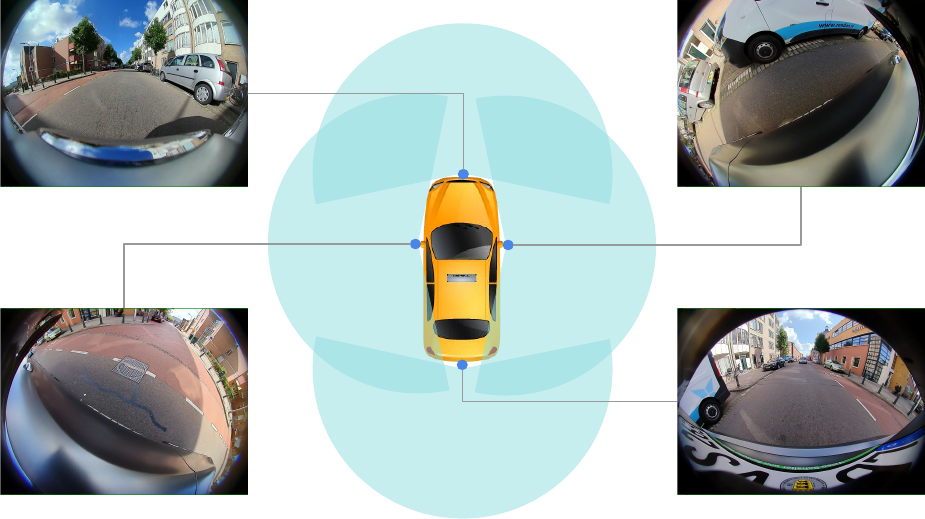}
   \caption{{\bfseries A four-fisheye SVS (Surround-View System)}. It provides a 360° coverage around the vehicle with overlapping zones between adjacent cameras.}
   \label{fig:svs}
\end{figure}

Camera-based \gls{svs} is one key component in \gls{adas} and autonomous driving. They are widely used in functions such as \gls{bev} image generation \cite{ipm, BEV_Lin_2012, Zhang2014ASV}, parking assistance \cite{holistic_parking, Zhang2018, Xiao_2023}, and 3D perception \cite{BEVFusion, CenterPoint}. A typical \gls{svs} consists of four wide-angle fisheye cameras arranged around the vehicle, providing a 360° coverage (\cref{fig:svs}).

Although current offline extrinsic calibration methods \cite{pattern_based_2012, Zhang2014ASV, pattern_based_2021, ZhangLin_2019, ZhangLin_2020_OECS, ZhangLin_2021_ROECS, LiJixiang2023} can provide accurate calibrations in their target fields around the vehicle, most of those fields are in short range (typically less than 5 meters, \cref{fig:related_work_examples}). This limitation is due to two main reasons: first, for pattern-based methods, the distance is limited to the physical size of the pattern as well as the calibration space; second, for photometric-based methods, since the quality of the synthesized \gls{bev} image drop sharply at longer distances, the calibration field is hence also limited.

\begin{figure}[thpb]
    \centering
    \subfloat[\centering Points clicking]{\includegraphics[height=.38\linewidth]{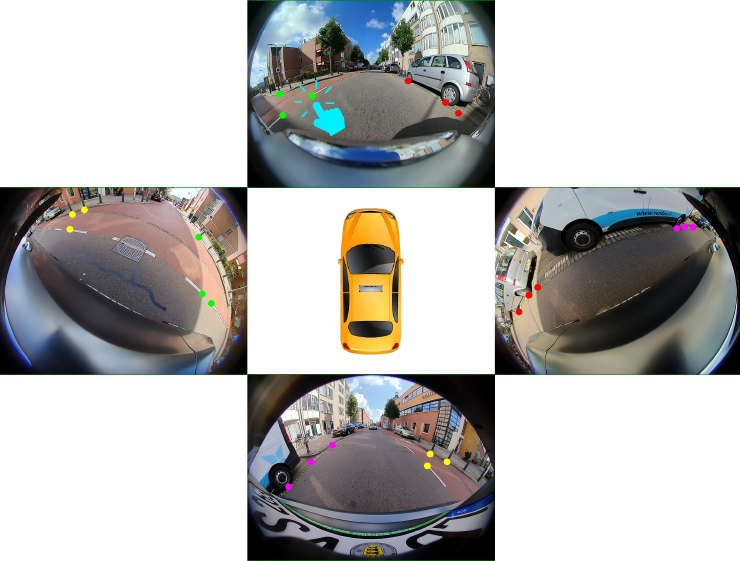}}
    \qquad
    \subfloat[\centering Generated \gls{bev} image]{\includegraphics[height=.38\linewidth]{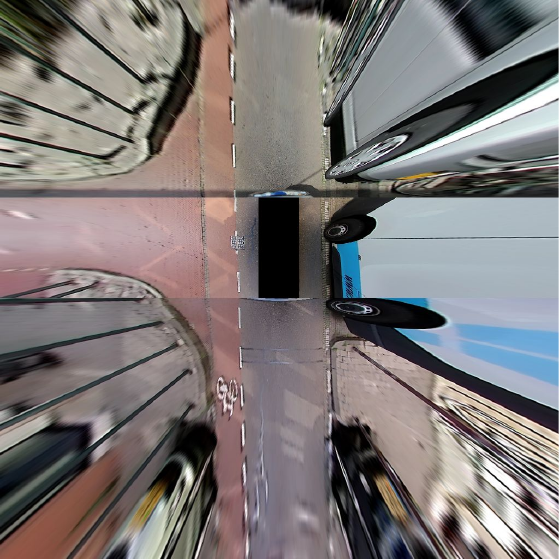}}
    \caption{{\bfseries Our proposed Click-Calib.} (a): User only needs to click a few points on the ground in the overlapping zones of adjacent \gls{svs} cameras (different point colors indicate overlapping zones between different pairs of cameras). Click-Calib then provides high-quality calibration results. (b): The generated \gls{bev} image using the calibration from Click-Calib.}
    \label{fig:click_calib}
\end{figure}

In this paper, we propose Click-Calib, a simple yet robust approach for the extrinsic calibration of \gls{svs} (\cref{fig:click_calib}). This method can be applied while the car is stationary or moving at low speeds (less than 30 km/h) on flat ground. Without requiring any special setup, the user only needs to select some keypoints on the ground in the overlapping zones of adjacent cameras. The calibration from Click-Calib maintains high accuracy at both short and long distances (greater than 10m), making it well-suited for long-range 3D perception tasks.

In summary, our contributions are threefold:

1) We propose Click-Calib, an extrinsic calibration method for \gls{svs} that requires no special setup or calibration patterns. Unlike other fisheye calibration approaches that necessitate image dewarping from fisheye to perspective, it optimizes calibration parameters directly from the raw fisheye images, thereby avoiding information loss.

2) We demonstrate that photometric error is not well-suited to reflect the quality of large-range \gls{bev} images. Instead, we introduce \gls{mde} as a more accurate metric.

3) Our approach is evaluated on three different vehicles. Compared to other offline calibration methods, Click-Calib shows significant improvements especially at long distances. Additional experiments also demonstrate its robustness to environmental uncertainty, such as variations in the height of each keypoint.


\section{Related work}
\label{sec:related_work}
The \gls{svs} is a particular type of multi-camera systems, which is composed of at least two cameras. In this section, we focus on previous work related to extrinsic calibration of multi-camera systems, especially \gls{svs}.

\subsection{Pattern-based methods}

\begin{figure}[thpb]
    \centering
    \begin{subfigure}{0.47\linewidth}
        \centering
        \includegraphics[height=\linewidth]{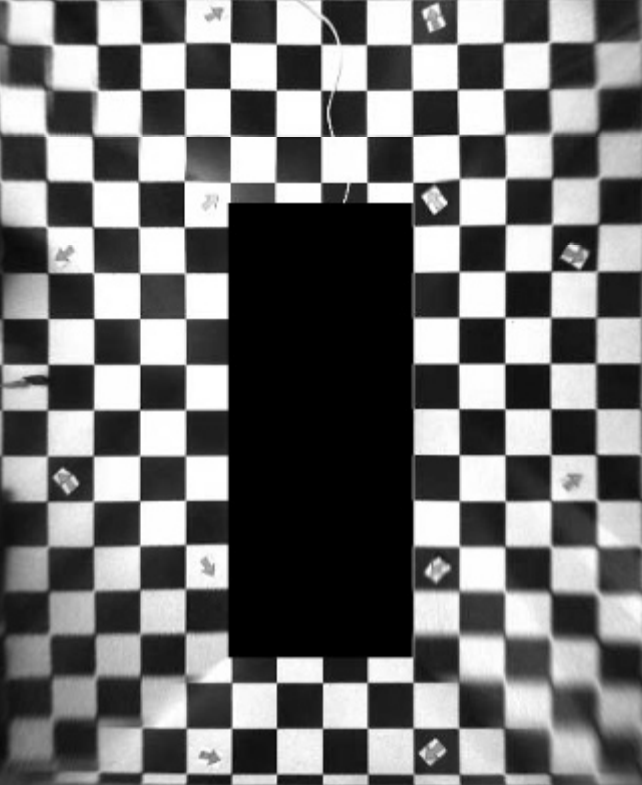}
        \caption{Pattern-based \cite{pattern_based_2012}}
        \label{fig:pattern_calib_example}
    \end{subfigure}
    \hfill
    \begin{subfigure}{0.47\linewidth}
        \centering
        \includegraphics[height=\linewidth]{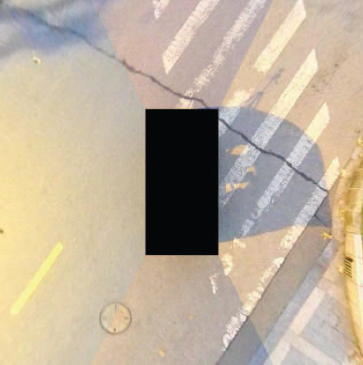}
        \caption{Photometric-based \cite{ZhangLin_2019}}
        \label{fig:photo_calib_example}
    \end{subfigure}
    \caption{{\bfseries Examples of pattern-based and photometric-based approaches.} The center dark box represents the ego vehicle. Both calibration methods focus on short range around the vehicle ($<$ 5 meters) due to the limitations of pattern size or \gls{bev} image quality.}
    \label{fig:related_work_examples}
\end{figure}

This category of methods is designed for offline calibration purposes. They are conducted while the car is stationary, using specific patterns (also known as calibration boards or targets) with known sizes to achieve high accuracy. Most of them also need a precise relative location between the vehicle and the patterns, which should be measured before the calibration. Consequently, a dedicated space is often required along with a time-consuming setup. Additionally, due to the physical size limitations of the patterns, they can only focus on short-range areas around the vehicle (typically 2-5 meters). A typical pattern-based calibration setup \cite{pattern_based_2012} is shown in \cref{fig:pattern_calib_example}.

In \cite{pattern_based_2003, pattern_based_2007}, the authors adopted a factorization-based method to calibrate the multi-camera system by placing patterns between adjacent cameras. Zhang \etal \cite{Zhang2014ASV} uses chessboard-like patterns placed in the common zones of adjacent cameras. Before performing calibration, they first apply fisheye lens distortion correction to obtain perspective images, which can result in information loss. To address the time-consuming setup of conventional pattern-based calibration, J. Lee and D. Lee \cite{pattern_based_2021} employ four randomly placed patterns to estimate the calibration by minizing both square-shaped errors and alignment errors. Although their method significantly reduces the setup time and effort, the calibration range is limited to about 2 meters from the vehicle.

\subsection{Feature-based methods}

Feature-based methods are typically designed for online calibration scenarios where the vehicle is in motion, tracking natural or man-made features to adjust camera poses in real-time. Nedevschi \etal in \cite{Online2007} estimated the vanishing point using parallel lane markings for stereovision calibration, assuming that the relative extrinsic parameters of the stereo cameras are known.  In \cite{Choi2018}, Choi \etal proposed a two-step approach to calibrate \gls{svs} by aligning lane markings across images from adjacent cameras. However, their method still assumes that lane markings are parallel, which limits its applicability.

In \cite{Koba2017}, Natroshvili \etal combined pattern-based and feature-based methods. The car is driven around the patterns placed on the ground, and the calibration is automatically estimated by detecting features on these patterns. Inspired by \gls{slam}, Carrera \etal \cite{Carrera2011} first built monocular feature maps while the robot made controlled movements, then matched and aligned those maps in 3D using invariant descriptors to determine the relative poses between multiple cameras. In \cite{Heng2013}, Heng \etal first built a 3D map using visual odometry then solved \gls{svs} extrinsics by optimizing camera-odometry transforms.

Although the calibration fields of most feature-based methods are not limited to the close range near the vehicle, they require additional information such as odometry \cite{Carrera2011, Heng2013} and can only be applied in specific scenarios \cite{Online2007, Choi2018} (\eg when the car is driving on a straight road with clearly painted lane markings).


\subsection{Photometric-based methods}

These approaches aim to optimize photometric errors after reprojecting the \gls{svs} images into a \gls{bev} image, making them suitable for both offline and online calibrations. They originate from the direct methods in \gls{slam} \cite{slam_direct_methods, slam_svo}, where dense image pixels are used for better intensity alignment. A typical photometric-based calibration setup \cite{ZhangLin_2019} is shown in \cref{fig:photo_calib_example}.

In 2019, Liu \etal \cite{ZhangLin_2019} first proposed a photometric-based calibration algorithm. Their method consists of two models, ground model and ground-camera model, both of which can correct the camera poses by minimizing the photometric errors of overlapping areas. Based on this work, Zhang \etal \cite{ZhangLin_2020_OECS} designed a novel model, the bi-camera model, to construct the photometric errors in adjacent camera images. In \cite{ZhangLin_2021_ROECS} they further refined the bi-camera model and used multiple frames rather than a single frame to build the overall error, so as to improve the system’s robustness.

Although \cite{ZhangLin_2019, ZhangLin_2020_OECS, ZhangLin_2021_ROECS} can achieve accurate calibrations, they have strict requirements on the initial extrinsic parameters due to the non-convex nature of the photometric error optimization process. Consequently, these methods are more suited for online correction than initial calibration. To address this limitation, Li \etal \cite{LiJixiang2023} proposed a coarse-to-fine solution to avoid falling into local optima. However, their method requires the front camera calibration to be known in advance and can only calibrate the other three cameras in \gls{svs}. Furthermore, the multi-stage random search strategy makes the approach slower.

Since photometric-based approaches \cite{ZhangLin_2019, ZhangLin_2020_OECS, ZhangLin_2021_ROECS, LiJixiang2023} directly optimize pixel-level alignment in \gls{bev} images, they often result in more accurate calibration compared to pattern-based methods. However, those methods are still limited to short distances around the vehicle due to their stringent requirements of \gls{bev} image quality (\cref{fig:photo_calib_example}). Additionally, the heavy computational load on dense pixels necessitates dewarping fisheye images to perspective images for acceleration, which results in information loss.

\section{Method}

\subsection{Notation and Terminology}
In this paper, we use $\mathbf{P}$ to denote a 3D point in space and $\mathbf{p}$ to denote a 2D point in an image (i.e., a pixel coordinate). Superscripts are used to indicate the coordinate system. For example, $\mathbf{p}^{C} = [u, v]^\top$ represents a pixel in the 2D image of camera $C$, and $\mathbf{P}^C = [X^C, Y^C, Z^C]^\top$ represents a 3D point in the camera coordinate system.


\subsection{Fisheye camera models}

Since the invention of fisheye cameras in 1906\cite{1906WoodFisheye}, their large field of view (typically $\geq 180^\circ$) has led to widespread use in surveillance, augmented reality, and especially in automotive applications\cite{WoodScape}. Unlike pinhole cameras, which map 3D points linearly to a 2D image, fisheye cameras produce images with significant radial distortion, particularly near the image borders. 

\begin{figure}[t]
  \centering
   \includegraphics[width=0.8\linewidth]{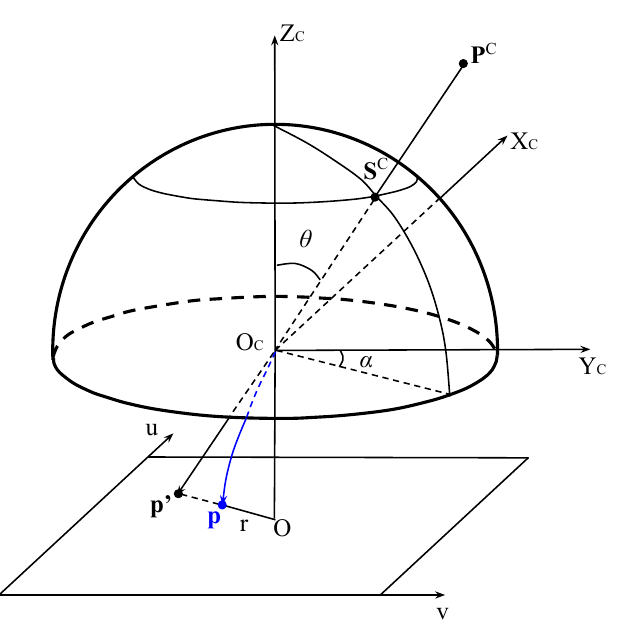}
   \caption{{\bfseries Fisheye camera model.} $\mathbf{P}^{C}$ is a 3D point in camera coordinate system. It intersects the unit sphere at $\mathbf{S}^C$ and is projected to $\mathbf{p}$ in the fisheye image (blue ray), whereas it would be projected to $\mathbf{p'}$ by a pinhole camera (black ray).} 
   \label{fig:fisheye_cam_model}
\end{figure}

To describe the strong radial distortion of fisheye lenses, various geometric models have been proposed (\!\cite{Hughes2010AccuracyOF, WoodScape, Kumar2022SurroundViewFC}). These models can be classified into four categories: classical geometric models, algebraic models, spherical models, and other models\cite{Kumar2022SurroundViewFC}. To facilitate the implementation of projection functions, we adopted an algebraic model in this paper. Specifically, following \cite{WoodScape}, we use a fourth-order polynomial

\begin{equation}
r = f(\theta) = a_1 \theta + a_2 \theta^2 + a_3 \theta^3 + a_4 \theta^4
\end{equation}

\noindent where $\theta$ denotes the incident angle and $r$ represents image radius in pixels. The coefficients $a_1$ to $a_4$ are distortion parameters obtained from intrinsic calibration. \cref{fig:fisheye_cam_model} illustrates the fisheye projection and compares it with the pinhole camera model.

Similar to the projection process, which maps a 3D point to a 2D image pixel, the reprojection (also known as unprojection) process of a camera is defined as the inverse operation, mapping a 2D image pixel back to a 3D point. For fisheye cameras, this involves calculating $\theta$ from $r$

\begin{equation}
\theta = f^{-1}(r)
\end{equation}

The analytical solution of fourth-order polynomial equation is complex\cite{quartic_equation}, therefore numerical approaches such as the Newton-Raphson \cite{gil2007numerical} method are often used in practice. The resolved $\theta$ can only provide a ray direction, as it is impossible to recover depth information during the 2D to 3D mapping. A straightforward representation of the reprojected ray is its intersection $\mathbf{S}^C = [X_{s}^C, Y_{s}^C, Z_{s}^C]^\top$ with the unit sphere (\cref{fig:fisheye_cam_model}), thus

\begin{equation}
\begin{aligned}
X_{s}^C &= \sin(\theta) \cos(\alpha) \vphantom{\frac{1}{\tan(\theta)}}, \\
Y_{s}^C &= \sin(\theta) \sin(\alpha) \vphantom{\frac{1}{\tan(\theta)}}, \\
Z_{s}^C &= \cos(\theta)
\label{eq:unit_sphere}
\end{aligned}
\end{equation}

\noindent where 

\begin{equation}
\alpha = \arctan\left(\frac{v-v_0}{u-u_0}\right)
\end{equation}

\noindent and 

\begin{equation}
r = \sqrt{(u-u_0)^2 + (v-v_0)^2}
\end{equation}

\noindent where $(u_0, v_0)$ denotes the principal point in pixel coordinates.

With \cref{eq:unit_sphere}, any 3D point $\mathbf{P}^{C}$ can be expressed as:

\begin{equation}
[X^C, Y^C, Z^C]^\top = \lambda\ \cdot [X_{s}^C, Y_{s}^C, Z_{s}^C]^\top
\label{eq:unprojection}
\end{equation}

\noindent where the scaling factor $\lambda$ is the depth (i.e., distance from the center of projection $O_C$) of point $\mathbf{P}^{C}$.

\subsection{Camera-vehicle projection}
\label{sec:cam2veh_proj}

The aforementioned fisheye camera model describes the transformation between a 2D fisheye image and the 3D fisheye coordinate system $C$. To obtain the 3D point coordinate in the vehicle cooridnate system $V$, a camera-vehicle projection is discussed here.

To simplify linear transformations, homogeneous coordinates are used. Thus, 2D $\mathbf{p}^{C}$ and 3D $\mathbf{P}^C$ are extended to $[u, v, 1]^\top$ and $[X^C, Y^C, Z^C, 1]^\top$, respectively. The homogeneous transformation matrix from $V$ to $C$, also known as the extrinsic matrix, is given by:

\begin{equation}
\mathbf{T}_{CV} = \begin{bmatrix}
r_{11} & r_{12} & r_{13} & t_x \\
r_{21} & r_{22} & r_{23} & t_y \\
r_{31} & r_{32} & r_{33} & t_z \\
0 & 0 & 0 & 1
\end{bmatrix}
\end{equation}

\noindent In this matrix, the vector $\mathbf{t}=[t_x, t_y, t_z]^\top$ represents the translation, describing the position of $C$ in $V$. And $r_{ij}$ represents the elements of the rotation matrix $\mathbf{R}$, which describes the orientation of $C$ relative to $V$. $\mathbf{R}$ can be calculated from quaternion $q_{CV}=[w, x, y, z]$:

\begin{equation}
\resizebox{0.9\linewidth}{!}{$
\mathbf{R} = 
\begin{bmatrix}
1 - 2y^2 - 2z^2 & 2xy - 2wz & 2xz + 2wy \\
2xy + 2wz & 1 - 2x^2 - 2z^2 & 2yz - 2wx \\
2xz - 2wy & 2yz + 2wx & 1 - 2x^2 - 2y^2
\end{bmatrix}
$}
\end{equation}

\noindent with the constraint which reduces the rotation degrees of freedom to three:

\begin{equation}
  w^2 + x^2 + y^2 + z^2 = 1 
\end{equation}

Using $\mathbf{T}_{CV}$, for a point $\mathbf{P}^V$ in the vehicle coordinate system, its corresponding point $\mathbf{P}^{C_i}$ in the camera coordinate $C$ is given by:

\begin{equation}
  \mathbf{P}^{C} = \mathbf{T}_{CV} \cdot \mathbf{P}^V
  \label{eq:veh2cam}
\end{equation}

Similarly, $\mathbf{P}^V$ can also be calculated from $\mathbf{P}^{C}$:

\begin{equation}
  \mathbf{P}^V = \mathbf{T}_{CV}^{-1} \cdot \mathbf{P}^{C}
  \label{eq:cam2veh}
\end{equation}

Combining \cref{eq:unprojection} and \cref{eq:cam2veh}, 

\begin{equation}
  \mathbf{P}^V = \mathbf{T}_{CV}^{-1} \cdot [\lambda X_{s}^C, \lambda Y_{s}^C, \lambda Z_{s}^C, 1]^\top
  \label{eq:unitcam2veh}
\end{equation}

Due to the presence of the scaling factor $\lambda$, $\mathbf{P}^V$ only determines a ray. However, since this paper considers ground points only, for which

\begin{equation}
  \mathbf{P}_{g}^V = [X_{g}^V, Y_{g}^V, 0]
  \label{eq:img2vehground}
\end{equation}

\noindent where $\mathbf{P}_{g}^V$ is the ground point in $V$. Here, $Z_{g}^V = 0$ because the origin of the vehicle coordinate is located on the ground. With this constraint, the scaling factor $\lambda$ can be uniquely determined, and $\mathbf{P}_{g}^V$ can then be calculated from \cref{eq:unitcam2veh}.



\subsection{Optimization}
\label{sec:optimization}

The optimization goal is to determine the pose of each camera, which consists of 6 parameters: three translations in vector $\mathbf{t}=[t_x, t_y, t_z]^\top$ and three rotations determined by the quaternion $q_{CV}=[w, x, y, z]$. As shown in \cref{fig:svs}, the \gls{svs} comprises four cameras $C_1$, $C_2$, $C_3$ and $C_4$. For a pair of adjacent cameras $C_i$ and $C_j$, if a ground point $\mathbf{P}_{g}^V$ is visible in both of them, then the reprojection distance error from the two cameras is:

\begin{equation}
   \epsilon(\mathbf{P}_{g}^V)  = \| G(\mathbf{p}^{C_i}) - G(\mathbf{p}^{C_j}) \|_2
   \label{eq:reproj_dist_error}
\end{equation}

\noindent where $\| \cdot \|_2$ denotes the Euclidean norm, $G$ is the ground reprojection function determined by \cref{eq:unitcam2veh} and \cref{eq:img2vehground}, $\mathbf{p}^{C_i}$ and $\mathbf{p}^{C_j}$ are pixel coordinates of $\mathbf{P}_{g}^V$ in $C_i$ and $C_j$, respectively.


Then the best estimate of the \gls{svs} calibration can be obtained by minimizing the following objective function:

\begin{equation}
  \begin{aligned}[b]
    J = {\sum_ {k=1}^{N} \epsilon(\mathbf{P}_{g, k}^V})
    \end{aligned}
    \label{eq:objective_func}
\end{equation}

\noindent where $N$ is the total number of selected keypoints from all pairs of adjacent cameras.

We employ the \gls{bfgs} algorithm \cite{BFGS_book} as the solver because of its efficiency in nonlinear optimization. However, since the problem is non-convex, iterative methods like BFGS can easily become trapped in local minima. To address this issue, Click-Calib requires a reasonable initial value, particularly for the rotation parameters. In practice, this initial value can be easily obtained from the nominal pose of each camera, or through manual adjustment of the \gls{bev} image.

\subsection{Scale ambiguity}

Although the proposed approach requires that the number of selected ground points exceeds the number of unknown parameters, scale ambiguity still exists (\cref{{fig:scale_ambiguity}}). Intuitively, if the world scale is reduced, the distance error will correspondingly decrease. Consequently, if dimensional constraints are not introduced during the optimization process, minimizing the distance error will ultimately result in a world scale approaching zero, which is not physically meaningful.

To resolve this problem, one of the three translation parameters needs to be fixed during the optimization process. In practice, the height of each camera is chosen as the fixed parameter because it is easy to measure.

\begin{figure}[t]
  \centering
   \includegraphics[width=0.95\linewidth]{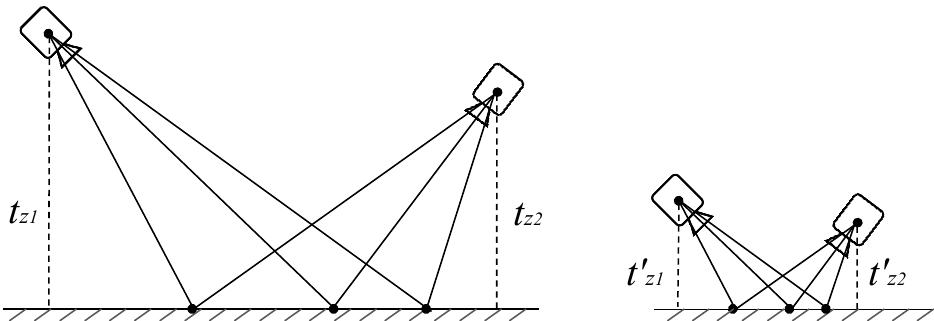}
   \caption{{\bfseries Scale ambiguity.} Left: larger world scale; Right: smaller world scale. Both setups yield identical camera images. Therefore, even with multiple cameras, the real-world scale (i.e., camera heights $t_{z1}$, $t_{z2}$ or $t'_{z1}$, $t'_{z2}$) cannot be determined solely through the projected ground points in the camera.}
   \label{fig:scale_ambiguity}
\end{figure}

\section{Experiments}

\subsection{Experimental setup}

The proposed Click-Calib is tested on two datasets: our in-house dataset (collected by two cars, referred to as Car 1 and Car 2 in the following sections) and the public WoodScape dataset\cite{WoodScape} (collected by one car). All three cars are equipped with four fisheye \gls{svs} cameras, providing 360° coverage around the ego vehicle with overlapping zones, as illustrated in \cref{fig:svs}. The image resolution in our dataset is is 1280 × 800, while in the WoodScape dataset it is 1280 × 966. To avoid image desynchronization issues at high speed, we only consider frames where the car's speed is less than 30 km/h. The collected images cover three critical scenarios—indoor parking, outdoor parking, and city driving—to demonstrate the robustness of the proposed approach. 

For each vehicle, the frame(s) used for calibration are referred to as the calibration set (similar to the training set in machine learning), and those used for evaluation as the test set. The fisheye images in both calibration and test sets are randomly selected from a consecutive image sequence. For cars in our dataset, we first calibrate the \gls{svs} cameras using a pattern-based conventional approach, which serves as the baseline for comparison with Click-Calib. For the car used in WoodScape dataset, we use the provided calibration as the baseline.

During the calibration process of Click-Calib, the keypoints are manually selected in each \gls{svs} image. To ensure the optimized calibration maintains high accuracy at different distances, a minimum of 10 keypoints are required in each overlapping zone.

For quantitative results, we generated \gls{bev} images using \gls{ipm} technique\cite{ipm, BEV_Lin_2012}. \gls{ipm} is widely employed in self-driving applications such as lane and parking detection~\cite{6957662, holistic_parking, Zhang2018, Wu2020, holistic_parking_patent, zhang2024}. Assuming that the world is flat, it generates a \gls{bev} image by projecting camera images onto the ground. To clearly demonstrate the quality of calibration's reprojection, we overlay all pixels reprojected from each camera. This visualization method provides an intuitive way to assess the calibration accuracy, as poor calibration will result in a severe "ghosting" effect in the \gls{bev} image due to misalignment from different \gls{svs} cameras. Examples of poor and good \gls{bev} images are shown in (c)-(d) of \ref{fig:metric}.

\subsection{Metrics}

To assess calibration quality, recent work on \gls{svs} calibration ~\cite{ZhangLin_2019, ZhangLin_2020_OECS, LiJixiang2023} uses photometric error (also known as photometric loss) as the metric. It measures the intensity differences of all pixels between two \gls{bev} images. The photometric error for a pair of adjacent cameras $C_i$ and $C_j$ is defined as

\begin{equation}
   \epsilon_{photo}  = \|\mathbf{I}_{C_i} - \mathbf{I}_{C_j} \|_2
   \label{eq:photo_error}
\end{equation}

\noindent where $\mathbf{I}_{C_i}$ and $\mathbf{I}_{C_j}$ are \gls{bev} images generated from cameras $C_i$ and $C_j$, respectively.

However, photometric error has two main limitations. First, \gls{svs} images are captured by different cameras with variations in illumination and exposure. These differences can cause high photometric error values even for well-aligned images. Second, large-range \gls{bev} images generated by \gls{ipm} often include objects above the ground, such as cars and walls, which cannot be properly aligned across different camera views. This misalignment also leads to significant photometric errors.

\begin{figure}[thpb]
    \centering
    \subfloat[\centering Front camera]{\includegraphics[width=.445\linewidth]{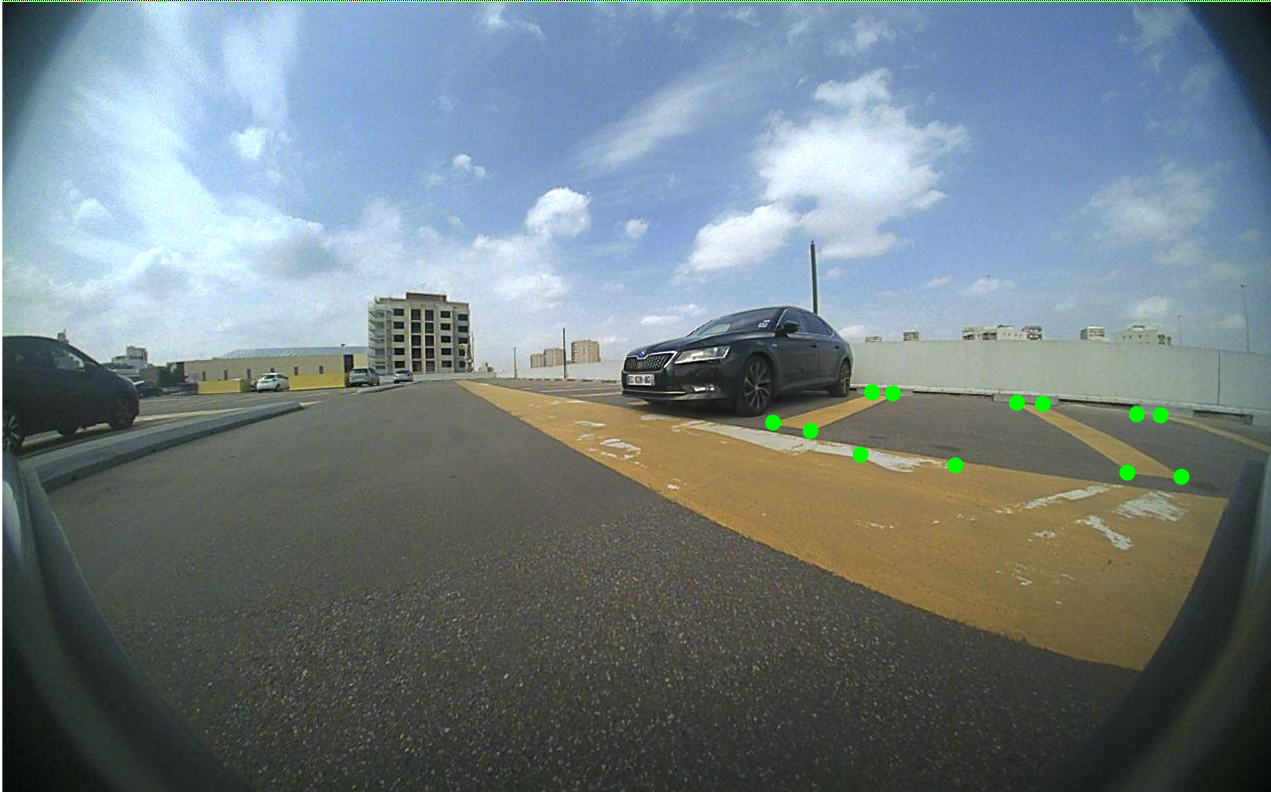}\label{fig:front} }
    \qquad
    \subfloat[\centering Right camera]{\includegraphics[width=.445\linewidth]{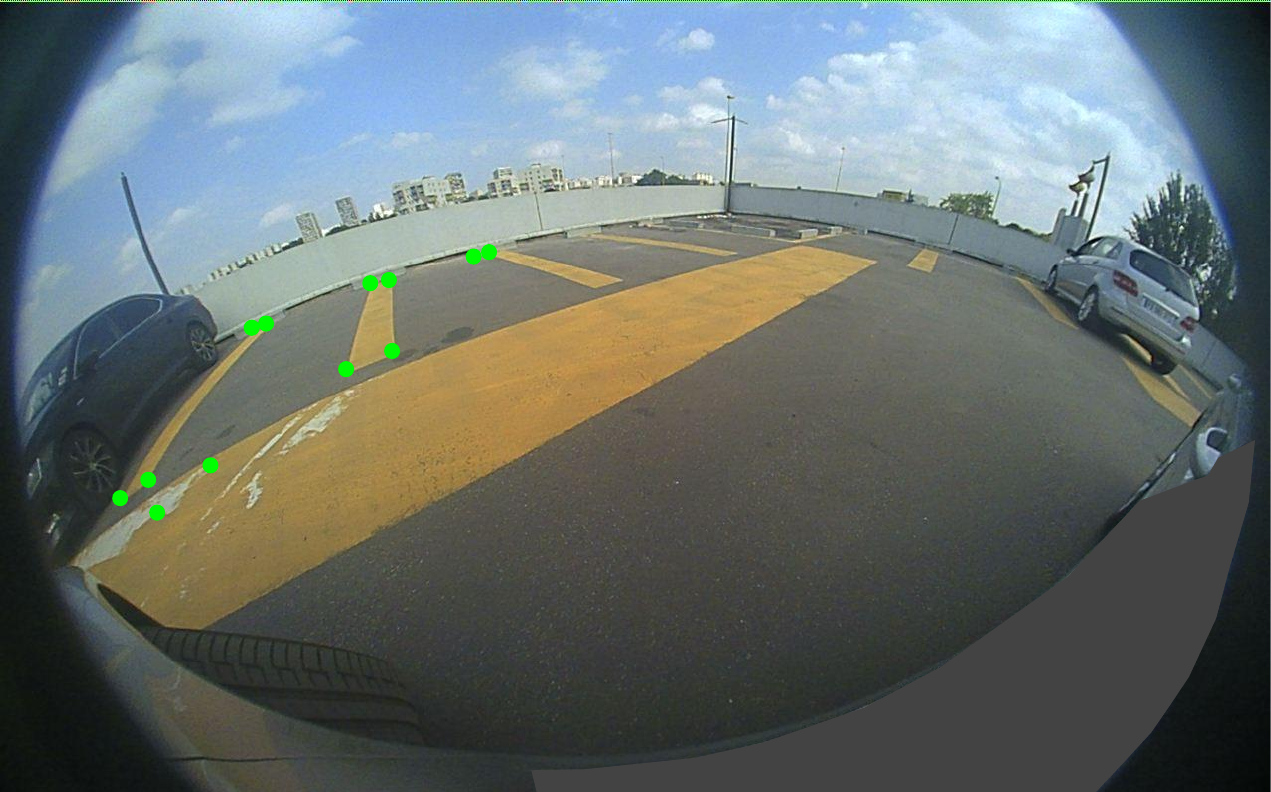}\label{fig:rear} }
    \qquad
    \subfloat[\centering BEV image (baseline)]{\includegraphics[width=.445\linewidth]{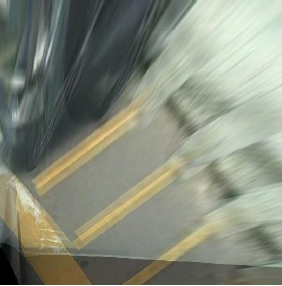}\label{fig:Left} }
    \qquad    
    \subfloat[\centering BEV image (ours)]{\includegraphics[width=.445\linewidth]{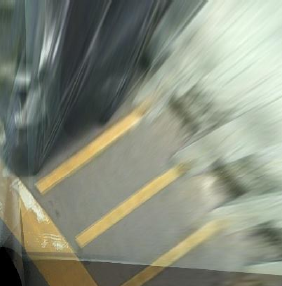}\label{fig:right} }
    \caption{{\bfseries Metrics comparison.} (a)-(b): \gls{svs} images (selected keypoints are marked in green). (c)-(d): \gls{bev} images for the overlapping zone. Mean distance error: 0.51m (baseline) vs. 0.21m (ours), \gls{bev} photometric error: 0.28 (baseline) vs. 0.31 (ours). Note that photometric error cannot accurately reflect \gls{bev} image quality due to variations in illumination and exposure among SVS cameras, as well as the presence of objects above the ground (\eg, cars, walls).}
    \label{fig:metric}
\end{figure}

To address the limitations of photometric errors, we employ a metric called Mean Distance Error (MDE), which aligns with our objective function described in \cref{sec:optimization}. Specifically, for each evaluation frame, we randomly select $M$ keypoints on the ground ($M$ is fixed to 20 in our experiments), then calculate the average reprojection distance error (\cref{eq:reproj_dist_error} and \cref{eq:objective_func}). Unlike photometric error, the proposed \gls{mde} is invariant to camera properties and non-flat objects, providing a fair assessment on the \gls{bev} image quality. A comparison of the photometric error and the \gls{mde} is shown in \cref{fig:metric}.

\subsection{Single-frame calibration}
\label{sec:single_frame_calib}

\begin{figure}[t]
  \centering
   \includegraphics[width=1\linewidth]{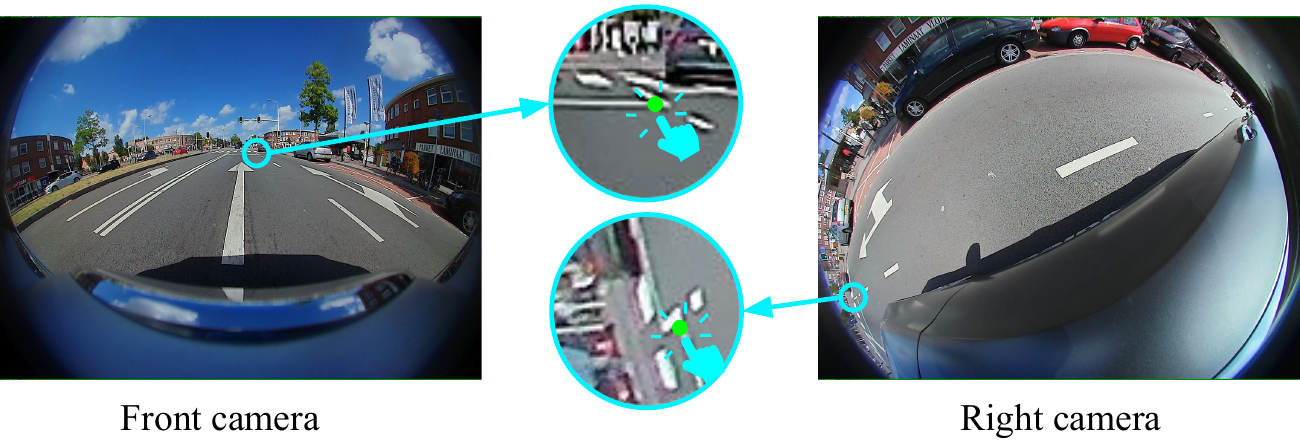}
   \caption{{\bfseries Distant keypoints example.} The selected keypoint is 15 meters from the ego vehicle, which enables Click-Calib to maintain high accuracy at greater distances.} 
   \label{fig:far_points}
\end{figure}

We first perform Click-Calib using only one frame as the calibration set. The results are listed in \cref{tab:single_frame}. For all three cars, our proposed approach surpasses the baseline in the \gls{mde} metric.  At shorter distance (i.e., areas closed to the ego vehicle), both the baseline and Click-Calib are accurate. However, at greater distances, especially those beyond 10 meters, the calibration from Click-Calib significantly outperforms the baseline. This is because Click-Calib allows the user to select keypoints at far distances (as long as they are visible in both adjacent cameras, see \cref{fig:far_points}), introducing more geometric constraints when solving for calibration. This feature makes Click-Calib particularly well-suited for recent \gls{bev}-based perception approaches, such as \cite{BEVFusion, CenterPoint, F-CVT}, which require high-accuracy calibration at long ranges.

\begin{table}[!htb]
  \centering
  {\small{
  \begin{tabular}{@{}lccccc@{}}
    \toprule
    Dataset and Method & 0-5m & 5-10m & $>$10m & Total\\ 
    \midrule
    \ \ \ Car 1 (baseline) &0.17 &0.34 &2.97 &1.31 \\
    \ \ \ \ \ \ Car 1 (ours) &0.08 &0.22 &2.21 &\textbf{0.95} \\
    \ \ \ Car 2 (baseline) &0.22 &0.39 &11.51 &4.45 \\
    \ \ \ \ \ \ Car 2 (ours) &0.22 &0.28 &2.06 &\textbf{0.93} \\
    \midrule
    WoodScape (baseline) &0.16 &0.50 &3.98 &1.56 \\
    \ \ WoodScape (ours) &0.14 &0.30 &2.53 &\textbf{0.99} \\
    \bottomrule
  \end{tabular}
  }}
  \caption{MDE (in meters) of single-frame calibration at different distances. Click-Calib outperforms the baseline on all three cars.}
  \label{tab:single_frame}
\end{table}

\begin{figure*}[h]
    \centering
    \subfloat
    {\includegraphics[width=.45\linewidth]{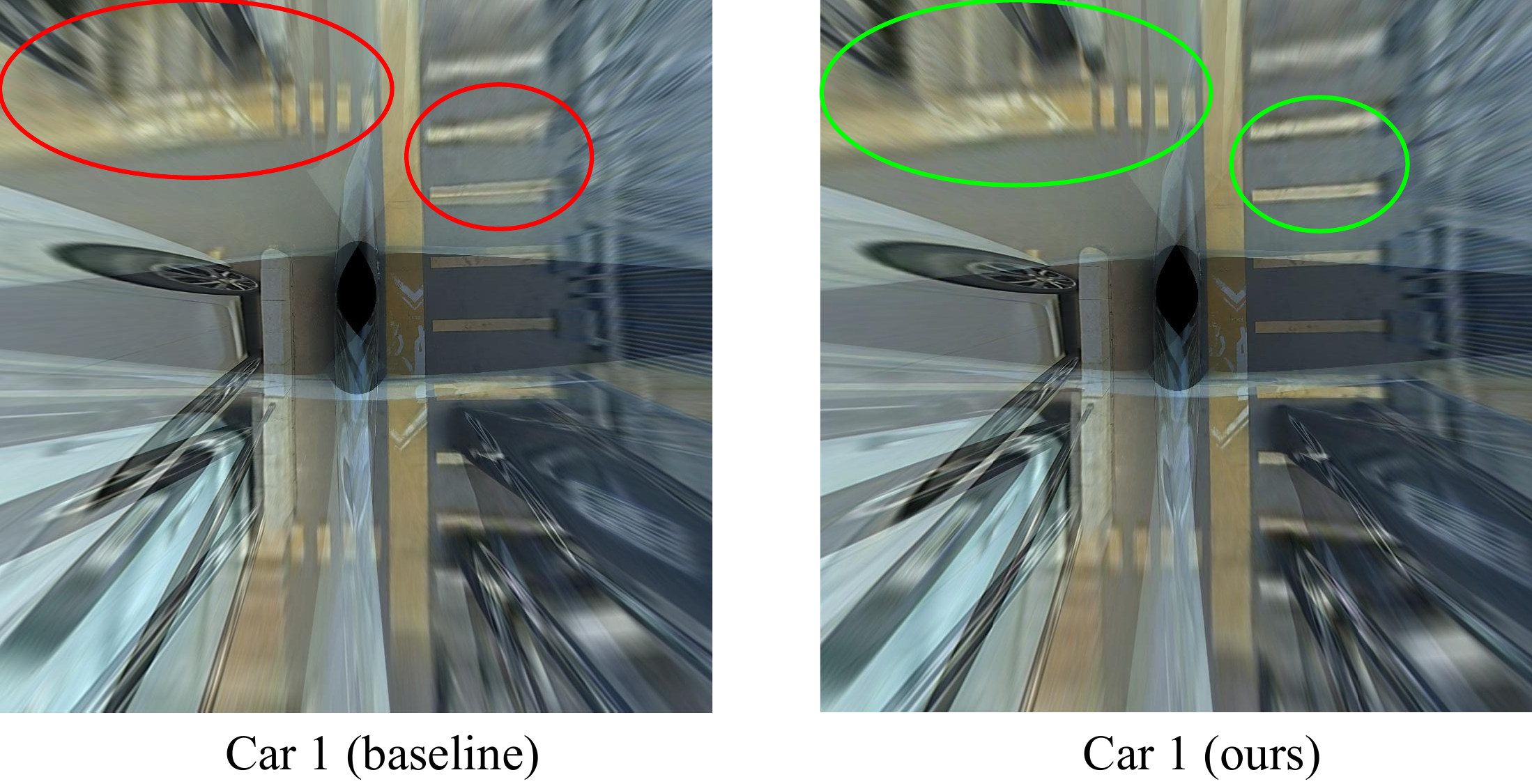}}
    \hspace{0.2cm}
    \qquad
    \subfloat
    {\includegraphics[width=.45\linewidth]{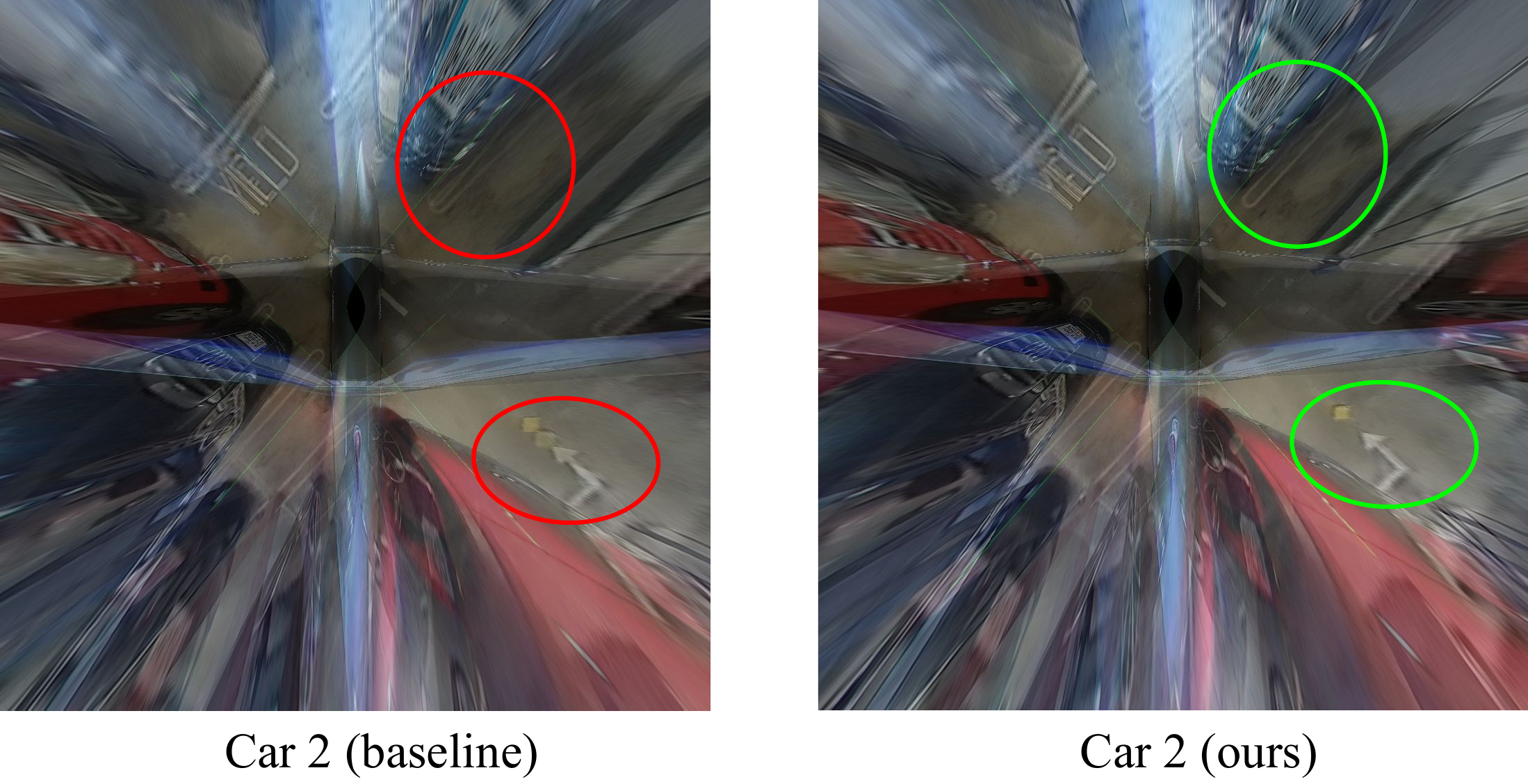}}
    \qquad
    \subfloat
    {\includegraphics[width=.45\linewidth]{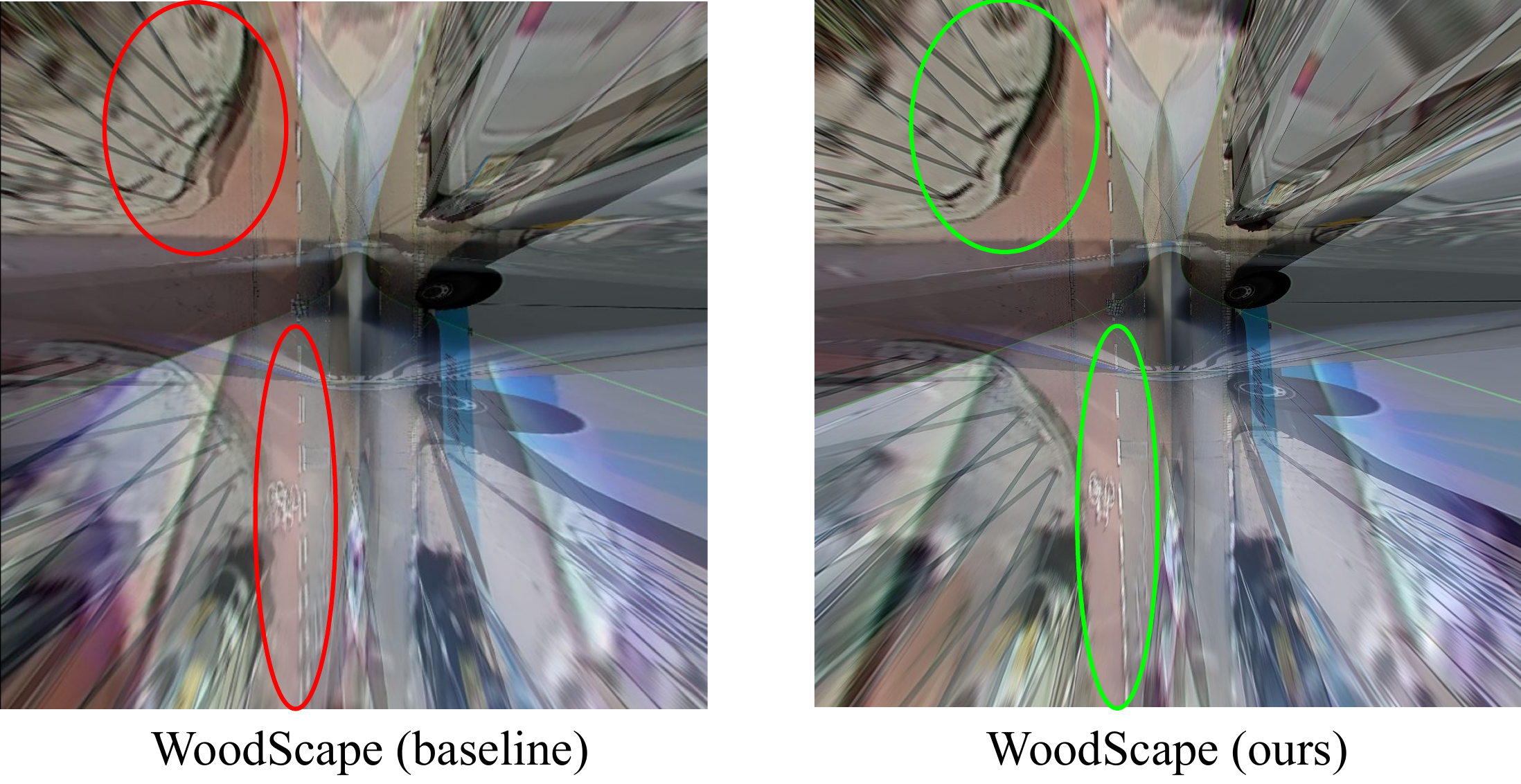}}
    \qquad
    \hspace{0.2cm}
    \subfloat
    {\includegraphics[width=.45\linewidth]{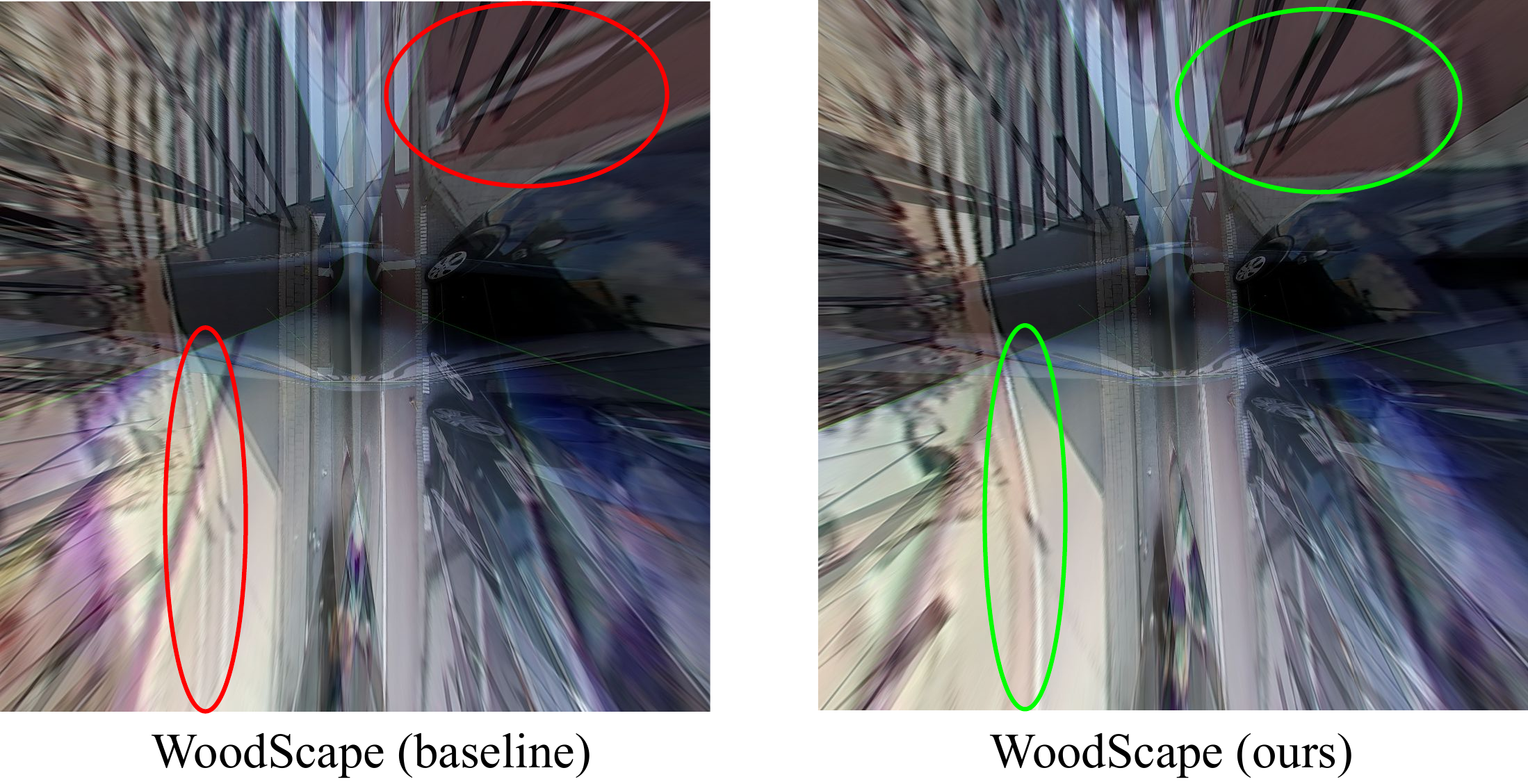}}
    \qquad
    \caption{{\bfseries Qualitative results.} In overlapped \gls{bev} images, the proposed Click-Calib provides much better alignment (green circles) than the baselines (red circles).}
    \label{fig:quali_res}
\end{figure*}

Some qualitative results are shown in \cref{fig:quali_res}. The generated \gls{bev} images cover a range of 25m × 25m around the vehicle. Compared to baselines, Click-Calib provides significantly better alignment between adjacent cameras, demonstrating its high accuracy.


\begin{table*}[!htb]
  \centering
  {\small{
  \begin{tabular}{@{}lcccccc@{}}
    \toprule
     \ Ground Type & $\Delta t_{x}$(max/mean) & $\Delta t_{y}$(max/mean) & $\Delta$roll(max/mean) & $\Delta$pitch(max/mean)  & $\Delta$yaw(max/mean) & MDE\\ 
    \midrule
    \ \ \ No noise &- &- &- &- &- &0.95m \\
    \ \ Slope noise & 0.05m / 0.02m & 0.05m / 0.03m & 0.11° / 0.07° & 0.08° / 0.05° & \textbf{0.92°} / 0.47° &0.84m\\
    Random noise & 0.06m / 0.03m & \textbf{0.11m} / 0.07m & 0.18° / 0.12° & 0.27° / 0.16° & 0.53° / 0.24° &0.99m\\
    \bottomrule
  \end{tabular}
  }}
  \caption{Robustness test results. Despite perturbations in keypoint height, Click-Calib maintains precise calibration. The mean and max values represent the average and maximum errors across the four cameras. The highest translation and angle errors are highlighted in bold.}
  \label{tab:Robustness}
\end{table*}

\subsection{Multiple-frame calibration}
Although the proposed approach can already provide high-quality calibration using only one frame, we also conducted calibration with multiple frames to mitigate the potential overfitting issue on a single frame.

This experiment is performed on Car 1. From a consecutive image sequence, we randomly selected $F$ frames (where $F$ ranges from 1 to 5) as the calibration set, and the test set remains the same as in \cref{sec:single_frame_calib}. The results are shown in \cref{tab:multi_frame}. The \gls{mde} significantly decreases with three frames and stabilizes when more than three frames are used. This improvement can be attributed to two main factors. First, additional frames provide more keypoints, offering broader coverage around the ego vehicle, which reduces the overfitting effect of using only one frame. Second, more frames also help smooth out the ground's unevenness, leading to more accurate calibration.


\begin{table}[!htb]
  \centering
  {\small{
  \begin{tabular}{@{}lcccccc@{}}
    \toprule
    Number of Frames & 1 & 2 & 3 & 4 & 5\\ 
    \midrule
    \ \ \ \ \ \ \ MDE (m) &0.95 &0.95 &0.69 &0.70 &0.72 \\
    \bottomrule
  \end{tabular}
  }}
  \caption{Multiple-frame calibration results.}
  \label{tab:multi_frame}
\end{table}

\subsection{Robustness test}

In reality, the ground is not perfectly flat, meaning that the assumption of $Z_{g}^V = 0$ in \cref{sec:cam2veh_proj} does not always hold. The height error $Z_{g}^V$ of keypoints can introduce inaccuracies in the optimized calibration. To quantify this error, one straightforward solution is to precisely measure the height of each keypoint. However, this process is time-consuming and requires expensive equipment. Therefore, we use simulations to estimate this error.

The \gls{iri} is the most commonly used index for measuring road roughness. It is defined as the accumulated vertical displacement of a standard reference vehicle relative to a flat road, measured over a given travel distance\cite{IRI_1, IRI_2}. The \gls{iri} value is usually expressed in meters per kilometer (m/km) or inches per mile (in/mi). For paved roads, \gls{iri} ranges between 1.5 to 6 m/km\cite{IRI_1}. We adopt the worst-case scenario 6 m/km as road roughness for our error estimation. The keypoints selection is limited to ±20m around the ego vehicle, therefore for each side of the ego vehicle the maximum variance in height is

\begin{equation}
   \Delta{Z_{g}^V}  = \frac{20}{1000} \cdot \ 6\,\text{m/km} = 0.12 \ \text{m}
\end{equation}

\begin{figure}[thpb]
    \centering
    \subfloat[\centering Scenario 1: slope noise]{\includegraphics[width=.8\linewidth]{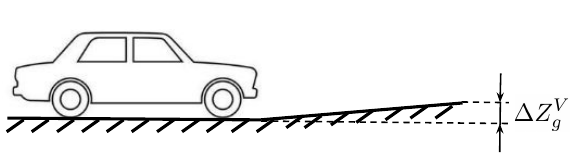}}
    \qquad
    \subfloat[\centering Scenario 2: random noise]{\includegraphics[width=.8\linewidth]{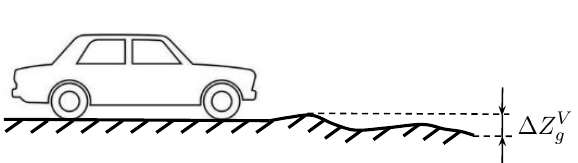}}
    \qquad
    \caption{{\bfseries Robustness test setup}. Only the front side of the ego vehicle is shown for simplicity.}
    \label{fig:robustness}
\end{figure}

This simulation is performed on Car 1 using single-frame calibration. We focus on two typical scenarios: the slope case and the random case. To simplify the analysis, we assume that the plane formed by the four wheels of the ego vehicle is perfectly horizontal. In the slope case, the ego vehicle is surrounded by slopes of height $\Delta{Z_{g}^V}$ on each side. In the random case, it is parked on a bumpy road, with the height of each point on the ground varying with random noise up to $\Delta{Z_{g}^V}$ (\cref{fig:robustness}).

The results of the robustness test are shown in \cref{tab:Robustness}. The simulated noises in ground point heights only introduces minor differences in calibrations. These differences can be considered an approximate upper bound of the calibration error from Click-Calib, as the primary source of error is from the heights of ground points. Interestingly, the \gls{mde} in the slope case is even smaller than in the no-noise case. We believe this is because the assumed slope partially matches the actual slope.

\section{Conclusions}

We proposed Click-Calib, a pattern-free extrinsic calibration approach for fisheye \gls{svs}. This method achieves accurate calibration with only some clicks on the ground in the overlapping zones of adjacent cameras. Compared to conventional pattern-based and recent photometric-based approaches, Click-Calib has three main advantages: (i) it is easy and fast to use without requiring special setup, (ii) it delivers high accuracy at both short and long distances (greater than 10 m), and (iii) it is robust to keypoint height noise. These features make it particularly well-suited for the recently prevalent \gls{bev}-based perception approaches.

\noindent {\bfseries Limitations and future work}. Although Click-Calib provides reliable calibration across all distances, it is subject to certain limitations. First, it is only effective when the vehicle is stationary or moving at low speeds (less than 30 km/h). Second, it requires manual clicking, which can be tedious for the user. Consequently, it is designed for offline calibration on small-batch vehicles rather than mass production. To transform it into a more general and fully automated method, our future work will focus on automating keypoints selection and extending keypoints from the ground-only to the entire 3D scene.










\end{document}